%% file: main.tex
\documentclass[10pt,twocolumn,letterpaper]{article}

\usepackage[pagenumbers]{cvpr}

\input{preamble}
\graphicspath{{./}{Arxiv/}}

\definecolor{cvprblue}{rgb}{0.21,0.49,0.74}
\usepackage[pagebackref,breaklinks,colorlinks,allcolors=cvprblue]{hyperref}

\def\confName{CVPR}
\def\confYear{2027}

\title{GLANCE-LT: Human Planning-Aware Modelling for \\  Long-Term Pedestrian Crossing Prediction}

\author{Leonardo Vinkestijn\\
University of Amsterdam\\
Amsterdam, The Netherlands\\
{\tt\small Leonardovinkestijn@gmail.com}
\and
Shaodi You\\
University of Amsterdam\\
Amsterdam, The Netherlands\\
}

\begin{document}

\twocolumn[{%
\renewcommand\twocolumn[1][]{#1}%
\maketitle
\centering
  
  \includegraphics[width=0.95\textwidth]{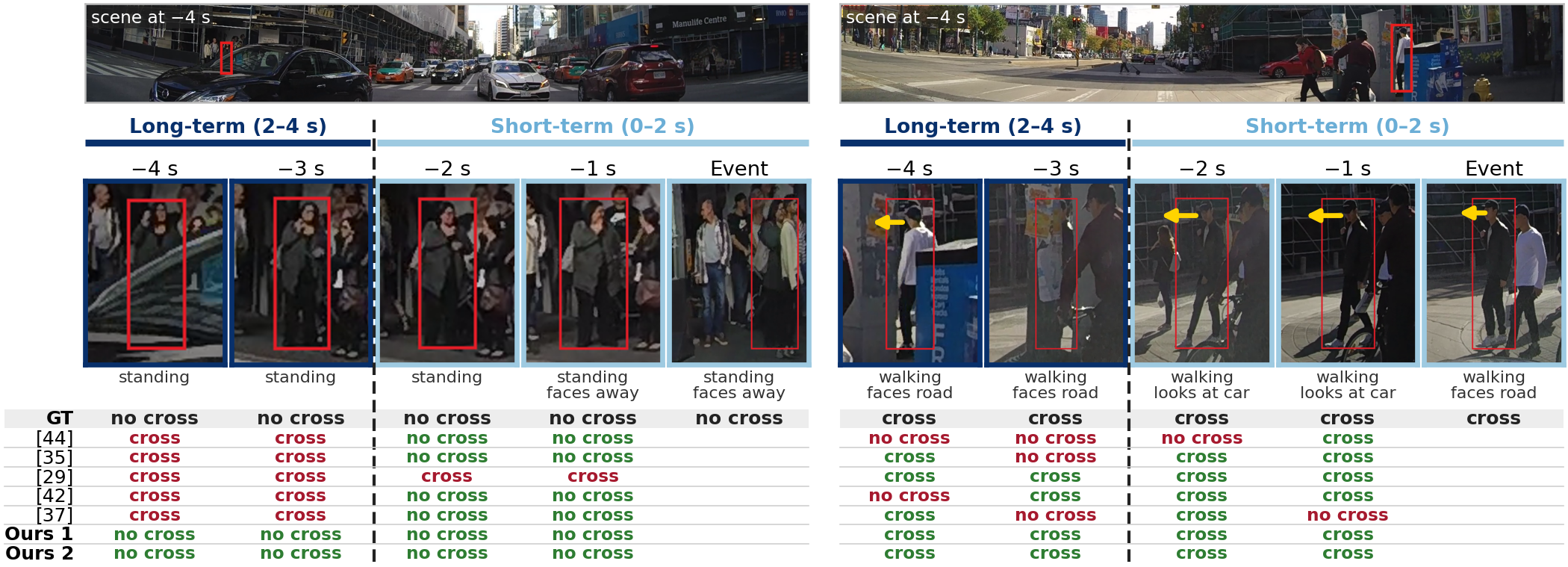}
  
      \captionof{figure}{
      Long-term ($>2$ s) vs short-term ($\leq2$ s) pedestrian crossing prediction on PIE~\cite{rasouli2019pie} datasets. Our methods model human behaviours related to long-term planning and thus make the correct predictions. Compared methods: TrEP (2023)~\cite{zhang2023trep}, VMI (2023)~\cite{sharma2023visualmotioninteraction}, TAMformer (2023)~\cite{osman2023tamformer}, PVIM (2024)~\cite{xu2024pvim} and ViCross~\cite{tian2026vicross}.
      Ours 1: VLM only, Ours 2: VLM + TrEP.
      }
  \label{fig:teaser}

}]

\begin{abstract}

Pedestrian Crossing Prediction (PCP) is a critical task for road safety and autonomous driving. Existing research primarily focuses on short-term prediction, typically within a two-second horizon. However, such a limited prediction horizon may not provide sufficient time for autonomous vehicles to anticipate potential risks and react safely. We therefore argue that long-term pedestrian crossing prediction (2+ seconds) is essential. Importantly, long-term prediction is not merely a straightforward extension of existing short-term approaches but introduces fundamentally different challenges. While short-term prediction primarily involves anticipating an imminent action, long-term prediction requires inferring a pedestrian's underlying intentions and future plans. To address this challenge, we propose GLANCE-LT, a novel prompt-based framework for long-term pedestrian crossing prediction. Our key insight is that understanding human planning requires reasoning over behavioral and contextual cues that reveal underlying intentions. Building on this insight, we develop a plug-and-play framework that integrates vision-language models (VLMs) with geometric reasoning to enable long-term crossing prediction. Extensive experiments across multiple benchmark datasets demonstrate the effectiveness of GLANCE-LT, achieving state-of-the-art performance with large performance gain in long-term pedestrian crossing prediction. Code will be released upon acceptance, and the supplementary material is available on Github\footnote{\url{https://github.com/Leonardovink/GLANCE-LT-Human-Planning-Aware-Modelling-for-Long-Term-Pedestrian-Crossing-Prediction}}.

\end{abstract}

\section{Introduction}

Pedestrian crossing prediction (PCP) is the task of predicting whether a pedestrian will cross the road in front of a vehicle, observed from the vehicle using video and vehicle dynamics. It is a critical task for road safety and autonomous driving~\cite{rasouli2019pie, kotseruba2021pcpa}. Existing research primarily focuses on short-term prediction ($\leq2$ s) \cite{kotseruba2021pcpa, xu2024pvim, zhang2023trep, tian2026vicross, azarmi2025pipvlfm}. However, road safety research shows that a vehicle needs at least 3 s for emergency manoeuvring~\cite{green2000stop,unece2008r13h,fmvss135,aashto2011green}. 
We therefore argue that long-term pedestrian crossing prediction (2+ seconds) is essential. 

Importantly, long-term prediction is not a straightforward extension of short-term prediction but introduces fundamentally different challenges. 
The fundamental difference is that short-term prediction primarily involves anticipating an imminent action, while long-term prediction requires inferring a pedestrian's underlying intentions and future plans.
Fig.~\ref{fig:teaser} shows this difference: most methods are right one to two seconds before the event, but wrong three to four seconds before it. Three seconds before the event the crosser on the right walks, but does not come closer to the road. What already shows is the plan to cross. The head is turned to the road, the light is red for the car and someone else is already crossing. Long-term prediction therefore requires inferring the intention of the pedestrian from how they plan the crossing and not from how they move.

To address this challenge, we propose GLANCE-LT, a novel prompt-based framework for long-term pedestrian crossing prediction. Our key insight is that understanding human planning requires reasoning over behavioural and contextual cues that reveal underlying intentions.
These cues come from how people prepare for crossing a road. When people walk, the head turns towards a new direction before the body does, and more than 90\% of pedestrians look at the approaching car before they cross~\cite{Hollands2002, GRASSO1998115, Rasouli2017agreeing}. For example, our analysis of the PIE dataset~\cite{rasouli2019pie} confirms this four seconds before the event: 73\% of the crossers have their head turned to the road against 37\% of the non-crossers, while moving towards the road differs only from 38\% to 30\%. We therefore developed seven cues in four groups for the model to check. Person cues describe what the pedestrian does (head and gaze, waiting, movement), scene cues where they are (crossing place, traffic light), the social cue whether others cross (other people), and the vehicle cue how the car moves (ego-vehicle dynamics). 

Building on these cues, we develop a plug-and-play framework that integrates vision-language models (VLMs) with geometric reasoning to enable long-term crossing prediction. We guide the VLM with these cues, not as rules but as evidence that the model should look for. We then adapt the VLM to dashcam footage with LoRA~\cite{hu2022lora} and combine its score with that of a geometric model, such as PVIM~\cite{xu2024pvim} or TrEP~\cite{zhang2023trep}, which still reads the motion that does appear. 

Extensive experiments on PIE~\cite{rasouli2019pie} and JAAD~\cite{kotseruba2016jaad} demonstrate the effectiveness of GLANCE-LT, achieving state-of-the-art performance in long-term prediction. On PIE, GLANCE-LT (VLM $+$ TrEP) gives the best AUC, F1, and accuracy at both 3 and 4 s, and at 3 s it raises them by 0.02, 0.05, and 0.03 over the best retrained baseline. Its semantic stream alone outperforms the previous VLM method ViCross~\cite{tian2026vicross} by 0.10 to 0.25 AUC at 3 s. On JAAD, without training on it, GLANCE-LT (VLM only) keeps an AUC of 0.63 at 4 s, where the best baseline reaches 0.56 and most stay below chance.

The contribution of this paper is fourfold:
\begin{itemize}
    \item To our knowledge, we give the first systematic study of long-term PCP as a research topic on its own.

    \item We provide a systematic analysis of human behaviour which indicates planning to cross in the long term.

    \item We propose \textbf{GLANCE-LT}, a two-stream framework for long-term PCP, which combines 
a geometric and a semantic stream. Especially, the semantic stream adopts human planning cues based on our systematic analysis.

    \item Extensive experiments validate the SoTA performance of the proposed method with a large performance gain.

\end{itemize}

\section{Related Work}

\subsection{Pedestrian crossing intention prediction}
\label{sec:related_pcp}

\noindent \textbf{Short-term prediction.} PCP is a relatively new task. Early work of Keller and Gavrila~\cite{keller2014ped} used vehicle-mounted stereo cameras to predict the path of a pedestrian and whether they would cross. 
The community mostly follows the benchmark of Kotseruba~\etal~\cite{kotseruba2021pcpa}, which observes 16 frames and then predicts whether a crossing will happen one to two seconds later on PIE~\cite{rasouli2019pie} and JAAD~\cite{kotseruba2016jaad}. 
Geometric and multimodal models fuse box trajectories, pose, ego speed, and local context~\cite{kotseruba2021pcpa, xie2025gtranspdm, 10208965, article, MCIP}. Zhang~\etal~\cite{zhang2023trep} propose TrEP, which models the box trajectory and the ego speed with a Transformer that also estimates its  own uncertainty. Xu~\etal~\cite{xu2024pvim} propose PVIM, which adds their pedestrian-vehicle spatial-context module that encodes relative position, bearing, distance, and ego speed. Azarmi~\etal~\cite{azarmi2025pipvlfm} prompted GPT-4V and LLaVA-NeXT, Uziel and Bialer optimise CLIP text embeddings~\cite{uziel2025clipcross} (ClipCross) and Munir~\etal~\cite{munir2025pedvlm} (PedVLM) pair a CLIP encoder with a T5 decoder. Ling~\etal and Mishra~\etal fine-tune models with chain-of-thought or question-answer supervision~\cite{Ling2026, Mishra2026}. Tian~\etal~\cite{tian2026vicross} propose ViCross, which fine-tunes Qwen2-VL-7B with variable-resolution patches to keep the target pedestrian readable. All of these methods are trained and tested at two seconds or less.

\noindent \textbf{Long-term prediction.} Predictions beyond two seconds exist but are under explored, both in number of publications and in performance. Fig.~\ref{fig:horizon_counts} shows the difference in the number of publications.
The experiments that do exist are very limited: long-term prediction is only an extra study next to a main result at one or two seconds.
Rasouli~\etal~\cite{Rasouli2020PedestrianAA} (SF-RNN) report their main result at 2 s and then sweep the time to event on PIE up to 3 s. 
Liu~\etal~\cite{liu2020spatiotemporal} (STIP) report their main result at 1 s and add 2 and 3 s on JAAD only. 
Ham~\etal~\cite{10208965} (CIPF) report their main result at 1 s and add a study at 2, 3, and 4 s on PIE only. Osman~\etal~\cite{osman2023tamformer} (TAMformer) and their successor~\cite{OSMAN2024104144}, which adds frame captions as a language modality to a multimodal transformer predict up to 4 s. However, both observe the whole track before the prediction (up to 4.5 s) instead of the 16 frames of the benchmark, and the code of the successor is not public.
Azarmi~\etal~\cite{azarmi2025pipnet} (PIP-Net) predict up to 4 s with their own protocol and their new multi-camera Urban-PIP dataset. 
Sharma~\etal~\cite{sharma2023visualmotioninteraction} (VMI) state that they predict 2.5 s ahead and only a figure shows their model at 4 s. These extra input requirements limit the applicability.
Where numbers are reported, the performance clearly drops as the horizon grows. For example, Ham~\etal~\cite{10208965} (CIPF) report an F1 drop from 0.84 at 1 s to 0.61 at 4 s. This indicates that long-term prediction is a challenging and understudied topic.

\noindent \textbf{Pedestrian behaviour before a crossing.}
Behaviour research from Hollands~\etal and Grasso~\etal~\cite{Hollands2002, GRASSO1998115} shows that when people walk, the head and eyes turn towards a new direction before the body does. At crosswalks without a signal, more than 90\% of pedestrians look at the approaching car before they cross~\cite{Rasouli2017agreeing}. These findings motivate the planning cues we analyse in Section~\ref{sec:analysis}.

\subsection{Datasets}

\label{sec:related_datasets}
PCP is mostly evaluated on two dashcam datasets. PIE, by Rasouli~\etal~\cite{rasouli2019pie}, which contains more than six hours of videos recorded in downtown Toronto at 30 fps with 1,842 annotated pedestrians, their crossing behaviour, the traffic lights, crosswalks and signs around them, and the ego-vehicle speed from the on-board diagnostics. JAAD, by Kotseruba~\etal~\cite{kotseruba2016jaad} contains 346 clips of 5 to 10 s recorded at several locations in North America and Europe with boxes for 2,786 pedestrians of whom 686 have behaviour labels.

Other datasets annotate pedestrians from a vehicle as well but the video samples are too short for long term predictions. TITAN, by Malla~\etal~\cite{malla2020titan}, aimed at forecasting trajectories rather than crossing. STIP, by Liu~\etal~\cite{liu2020spatiotemporal}, records busy intersections in eight cities in California and Michigan with three cameras and labels every pedestrian as crossing or not. Urban-PIP, by Azarmi~\etal~\cite{azarmi2025pipnet}, adds multi-camera views and is used with its own protocol.

\subsection{Vision-language models}

VLMs combine a vision encoder with a language model~\cite{pmlr-v139-radford21a, NEURIPS2023_6dcf277e, bai2025qwen25vl}. A strong general VLM on its own does not solve the task by itself. Huang~\etal~\cite{takesthewheel} show that zero-shot GPT-4V stay below specialized models on JAAD. Azarmi~\etal~\cite{azarmi2025pipvlfm} (PIP-VLFM) find that prompted VLMs show a strong dependence on ego speed given in text, and Tian~\etal~\cite{tian2026vicross} (ViCross) note that a distant pedestrian is only represented in a few image tokens. Luckily there are two simple input conventions that help with this. Shtedritski~\etal~\cite{10376723} and Cai~\etal~\cite{10657559} (ViP-LLaVA) show that a target marked with a bbox or a circle is understood without extra modules. Kim~\etal~\cite{imagegrid} (IG-VLM) show that a grid of frames lets an image VLM read a short video. We use both of these conventions and then adapt the model with LoRA~\cite{hu2022lora} and score its answers as a probability.

\section{Long-term PCP}
This section defines the task, shows the scientific and societal impact of long-term prediction and states the research question that separates long-term from short-term PCP.

\subsection{Task formulation}
\label{sec:task}
\noindent \textbf{Pedestrian crossing prediction}. As shown in Fig.~\ref{fig:terminology}, a sample x is a pedestrian observed from a vehicle-mounted dash camera. In common practice x is observed for 16 frames (0.53\,s at 30\,fps), and the task is to predict whether or not the pedestrian will cross the road at a future time, the time-to-event (TTE), which we also give in frames.

\noindent \textbf{Short- and long-term prediction.} 
Following our literature review (Fig.~\ref{fig:horizon_counts}) and without loss of generality, we define a TTE above 2 s as long-term.

\noindent \textbf{Scientific and Societal Impacts.}
Specifically, we argue that long-term PCP is a challenging new task because of its potential scientific impact: 1. the small number of existing studies, 2. the clear performance drop reported in the literature, and 3. the lack of systematic experimental validation.

Beyond that, it is clearly a task of societal impact. Pedestrians make up 23\% of the 1.19 million road deaths every year~\cite{world2023global}. In 2018 a self-driving test car in Tempe, Arizona detected a pedestrian 5.6 s before the impact, but only recognised the crash 1.2 s before it~\cite{ntsb2019tempe}. In real crashes emergency braking lowered pedestrian crashes by 27\%, but not on roads with a speed limit above 50 mph~\cite{CICCHINO2022106686}. With the braking that the UN regulation~\cite{unece2008r13h} and the US federal standard~\cite{fmvss135} require of every new car, a surprised driver needs 2.5 to 3.4 s to stop from 30 to 70 km/h (supplementary material). A vehicle therefore has to know two to four seconds ahead that a pedestrian will cross.
\definecolor{longc}{HTML}{08306B}
\definecolor{shortc}{HTML}{9ECAE1}
\definecolor{shorttxt}{HTML}{6BAED6}

\begin{figure}[t]
\centering
\begin{tikzpicture}[x=0.49cm, y=0.5cm, font=\scriptsize, >=stealth]

  \draw[->, thick] (0,0) -- (14.4,0);
  \foreach \x/\lab in {0/0, 1.6/16, 7.6/76}
    \draw (\x,0.12) -- (\x,-0.12) node[below=1pt] {\lab};
  \draw (13.6,0.12) -- (13.6,-0.12) node[below=1pt] {136 frames};

  \fill[blue!12] (0,0.25) rectangle (1.6,1.05);
  \draw (0,0.25) rectangle (1.6,1.05);
  \draw[<->] (0,1.45) -- (1.6,1.45);
  \node[above right, inner sep=1pt] at (0,1.55) {observation: 16 frames ($\approx$0.53\,s)};

  \foreach \x in {7.6, 13.6} \draw[densely dotted, gray] (\x,-0.75) -- (\x,0.45);
  \node at (9.1,0.85) {range of the crossing event};
    \draw[<->, shortc, line width=1.2pt] (1.6,-1.3) -- (7.6,-1.3);
  \node[shorttxt, font=\scriptsize\bfseries, align=center, anchor=north] at (4.6,-1.45)
    {TTE 0--60 (0--2\,s)\\short-term};
  \draw[<->, longc, line width=1.2pt] (7.6,-1.3) -- (13.6,-1.3);
  \node[longc, font=\scriptsize\bfseries, align=center, anchor=north] at (10.6,-1.45)
    {TTE 60--120 (2--4\,s)\\long-term (this paper)};

\end{tikzpicture}
\caption{Short- and long-term crossing prediction in this paper. From an observation of 16 consecutive frames (0.53 s), the common practice, the model predicts whether the pedestrian crosses a time-to-event (TTE) after the last observed frame. Following the literature in Fig.~\ref{fig:horizon_counts}, we call a TTE above 2 s long-term.}

\label{fig:terminology}
\end{figure}
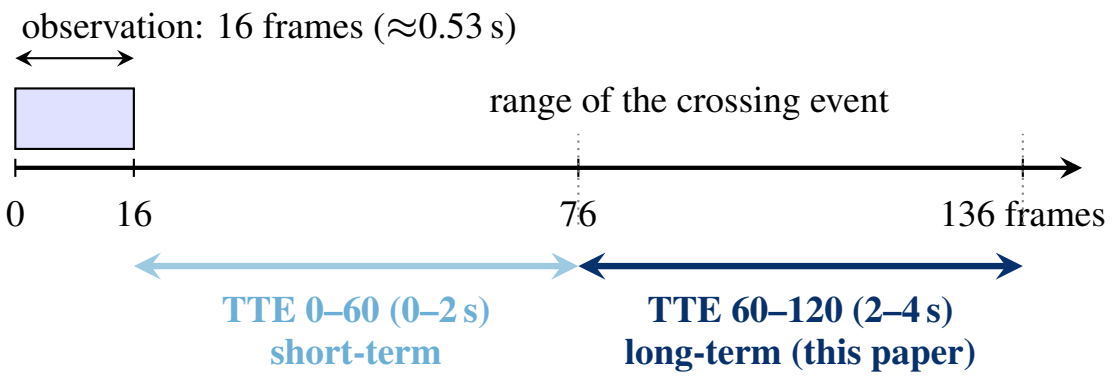

\definecolor{yr1}{HTML}{C6DBEF}
\definecolor{yr2}{HTML}{6BAED6}
\definecolor{yr3}{HTML}{2171B5}
\definecolor{yr4}{HTML}{08306B}
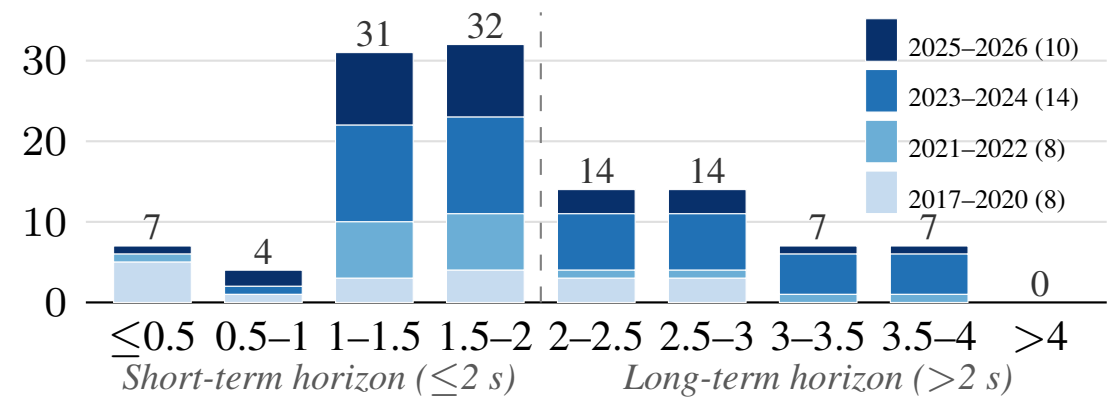
\begin{figure}[tb]
\centering
\begin{tikzpicture}
\begin{axis}[
  width=\linewidth, height=0.42\linewidth,
  ybar stacked, bar width=0.7,
  xmin=0.4, xmax=9.6, ymin=0, ymax=36, ytick={0,10,20,30},
  xtick={1,...,9},
  xticklabels={$\le$0.5, 0.5--1, 1--1.5, 1.5--2, 2--2.5, 2.5--3, 3--3.5, 3.5--4, $>$4},
  ticklabel style={font=\footnotesize},
  axis x line*=bottom, axis y line*=left,
  x axis line style={-}, y axis line style={draw=none},
  major tick style={draw=none},
  ymajorgrids, grid style={gray!25},
  enlarge x limits=false, clip=false,
    legend style={font=\tiny, draw=none, fill=none, at={(1.0,1.02)}, anchor=north east,
                legend cell align=left, row sep=-1pt},
  reverse legend,
]
\addplot[fill=yr1, draw=white, line width=0.2pt] coordinates {(1,5) (2,1) (3,3) (4,4) (5,3) (6,3) (7,0) (8,0) (9,0)};
\addlegendentry{2017--2020 (8)}
\addplot[fill=yr2, draw=white, line width=0.2pt] coordinates {(1,1) (2,0) (3,7) (4,7) (5,1) (6,1) (7,1) (8,1) (9,0)};
\addlegendentry{2021--2022 (8)}
\addplot[fill=yr3, draw=white, line width=0.2pt] coordinates {(1,0) (2,1) (3,12) (4,12) (5,7) (6,7) (7,5) (8,5) (9,0)};
\addlegendentry{2023--2024 (14)}
\addplot[fill=yr4, draw=white, line width=0.2pt] coordinates {(1,1) (2,2) (3,9) (4,9) (5,3) (6,3) (7,1) (8,1) (9,0)};
\addlegendentry{2025--2026 (10)}
\node[font=\scriptsize, text=black!80, anchor=south, inner sep=1pt] at (axis cs:1,7) {7};
\node[font=\scriptsize, text=black!80, anchor=south, inner sep=1pt] at (axis cs:2,4) {4};
\node[font=\scriptsize, text=black!80, anchor=south, inner sep=1pt] at (axis cs:3,31) {31};
\node[font=\scriptsize, text=black!80, anchor=south, inner sep=1pt] at (axis cs:4,32) {32};
\node[font=\scriptsize, text=black!80, anchor=south, inner sep=1pt] at (axis cs:5,14) {14};
\node[font=\scriptsize, text=black!80, anchor=south, inner sep=1pt] at (axis cs:6,14) {14};
\node[font=\scriptsize, text=black!80, anchor=south, inner sep=1pt] at (axis cs:7,7) {7};
\node[font=\scriptsize, text=black!80, anchor=south, inner sep=1pt] at (axis cs:8,7) {7};
\node[font=\scriptsize, text=black!80, anchor=south, inner sep=1pt] at (axis cs:9,0) {0};
\draw[dashed, gray] (axis cs:4.5,0) -- (axis cs:4.5,36);
\node[font=\scriptsize\itshape, text=black!65, anchor=north] at ([yshift=-8pt]axis cs:2.5,0) {Short-term horizon ($\le$2 s)};
\node[font=\scriptsize\itshape, text=black!65, anchor=north] at ([yshift=-8pt]axis cs:7,0) {Long-term horizon ($>$2 s)};
\end{axis}
\end{tikzpicture}
\caption{Number of PCP papers that test each prediction horizon. Long-term prediction ($>$2 s) is under-explored. Notice the long-term tests are not as extensive as short-term.}
\label{fig:horizon_counts}
\end{figure}

\noindent \textbf{The field tests one to two seconds.}
\label{sec:why}
The benchmark of Kotseruba~\etal~\cite{kotseruba2021pcpa} asks whether the pedestrian crosses 30 to 60 frames after the observation, and later geometric and vision-language methods follow it. Of the 40 papers in Fig.~\ref{fig:horizon_counts}, only 14 test beyond 2 s and only 7 reach 4 s. On the standard dashcam benchmarks, the range from two to four seconds has not been explored sufficiently and has no common protocol.

\subsection{Imminent action vs planning}
\label{sec:analysis}
\begin{figure}[t]
\centering
\includegraphics[width=\columnwidth]{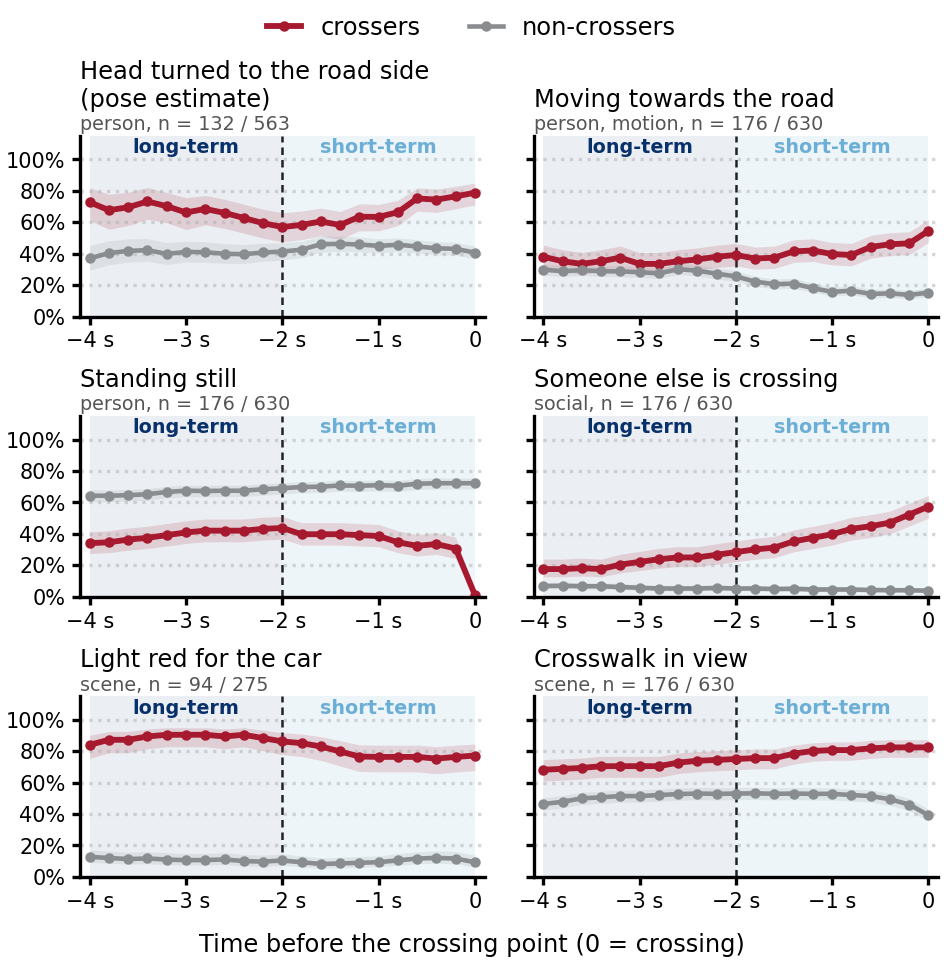}
\caption{Share of crossers and non-crossers that show each behaviour before the crossing. Some behaviours separate the two groups already in the long term, others only in the short term.}

\label{fig:timecourse}
\end{figure}

We further argue that long-term PCP is not a naive extension of short-term PCP but a genuinely different task: short-term prediction anticipates an action that has already begun, while long-term prediction anticipates an action that is still being planned.
\\\\\\\\\\
\noindent \textbf{Research Question.} We therefore propose the research question:

\textit{Can we model pedestrian behaviour based on observations and predict whether the pedestrian plans to cross the road in the long term?}

Take Fig.~\ref{fig:teaser} (right) as an example. Three seconds before crossing, the pedestrian walks but does not approach the road. However, the head turned to the road, the red light for vehicles, and another person already crossing indicate a plan to cross.
This observation is endorsed by the literature on human behaviour ~\cite{Hollands2002, GRASSO1998115, Rasouli2017agreeing} (Section~\ref{sec:related_pcp}).

We thoroughly analyse this behaviour on PIE~\cite{rasouli2019pie}. Fig.~\ref{fig:timecourse} follows six behaviours of crossers (red) and non-crossers (grey) over the TTE: three person cues (head turned to the road, moving towards the road and standing still), one social cue (someone else is crossing) and two scene cues (the light is red for the car and a crosswalk is in view). 

These behaviours clearly indicate the possibility of crossing, and they differ between the short and the long term. For moving towards the road and for someone else crossing, the difference between crossers and non-crossers grows towards the event, and for moving towards the road it only opens up in the last two seconds (38\% against 30\% at 4 s). Head orientation and standing still already separate the two groups four seconds before the event: 73\% of the crossers have their head turned to the road against 37\% of the non-crossers (Fig.~\ref{fig:timecourse}), and 34\% stand still against 64\%. Two to four seconds before the event, motion, which geometric models read well, therefore looks almost identical for crossers and non-crossers, while planning cues such as the head turned to the road, standing still, the traffic light, and other people crossing already differ. A long-term model has to read the pedestrian and the scene, and not only how they move.

\noindent \textbf{What this asks of a model.}
\label{sec:requirements}
The above analysis requires three abilities for a long-term model. First, it has to read the person and not the motion, because head orientation and waiting already differ four seconds before the event, while movement does not yet (R1). Second, the model has to read the scene and the people around the pedestrian, because the crossing place, traffic light, and other crossers carry evidence (R2). Third, it has to be able to see the pedestrian at a resolution where a head turn is visible, because two to four seconds before the event some pedestrians are small and their pose is then often not detected (R3). The method section builds the model around these three requirements.

\section{GLANCE-LT}
\label{sec:method}
\begin{figure*}[t]
\centering
\includegraphics[width=0.90\textwidth]{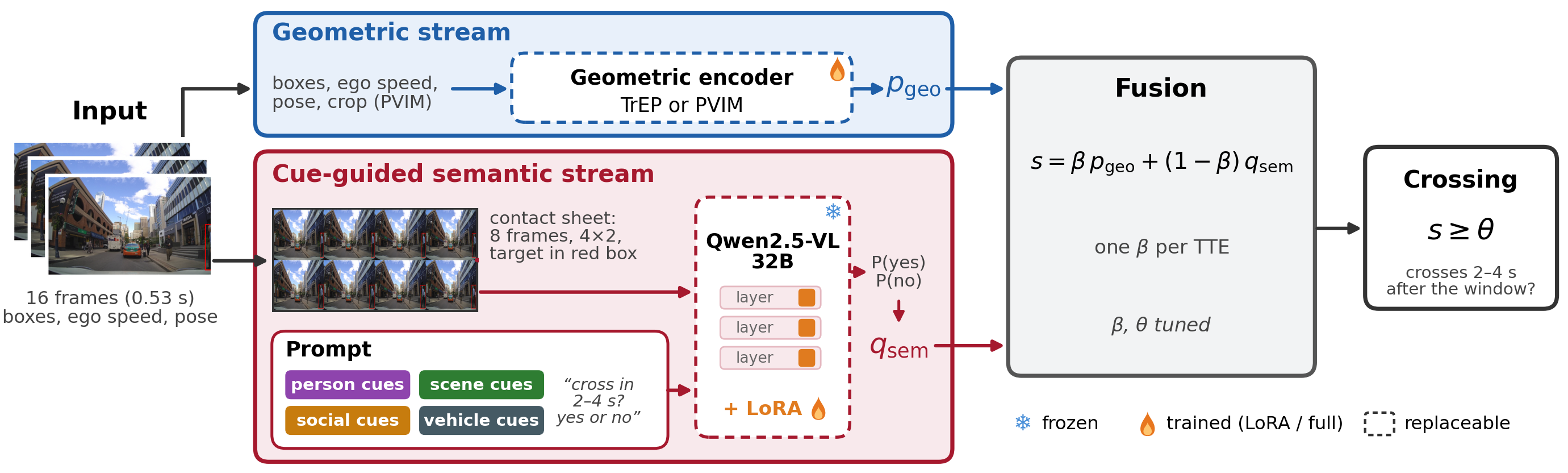}
\caption{Overview of GLANCE-LT. Both streams read the same 16-frame window. The fusion is a weighted sum of both scores, with $\beta$ and the threshold $\theta$ tuned per TTE (Section~\ref{sec:protocol}). Either stream can be replaced.}
\label{fig:method}
\end{figure*}

\subsection{Overview}
\label{sec:overview}
We propose \textbf{GLANCE-LT}, a novel prompt-based framework for long-term pedestrian crossing prediction.
Following Section~\ref{sec:analysis}, our key insight is that understanding human planning requires reasoning over behavioural and contextual cues that reveal underlying intentions. We therefore build a plug-and-play framework in which two streams read the same 16-frame window, the default in common practice, and a fusion step combines their scores (Fig.~\ref{fig:method}).
\begin{itemize}[leftmargin=*, itemsep=1pt, topsep=2pt]
    \item The \emph{semantic stream} is a VLM that reads the frames, guided by the cues of Section~\ref{sec:analysis}, and outputs $q_{\text{sem}}$.
    \item The \emph{geometric stream} reads where the pedestrian is and how the pedestrian and vehicle move extracted from the boxes and the ego speed, which outputs $p_{\text{geo}}$
    \item The \emph{fusion} combines both probabilities into one score $s = \beta p_{\text{geo}} + (1-\beta)q_{\text{sem}}$.
\end{itemize}

The semantic stream is built around the requirements of Section~\ref{sec:requirements}. Its prompt asks for the person (R1) and for the scene and the people around them (R2) (Section~\ref{sec:semantic}). A red bbox marks the target in every frame so that the model also finds small pedestrians (R3), although the resolution of the contact sheet limits how small a pedestrian can be while a head turn stays visible (Section~\ref{sec:limitations}). 

The streams are plug-and-play: the fusion only needs a probability from each stream, so either backbone can be replaced without changing the rest. We use Qwen2.5-VL-32B-Instruct~\cite{bai2025qwen25vl} as the semantic stream and TrEP~\cite{zhang2023trep} or PVIM~\cite{xu2024pvim} as the geometric stream. We call the semantic stream alone GLANCE-LT (VLM only) and its fusion with TrEP~\cite{zhang2023trep} or PVIM~\cite{xu2024pvim}, GLANCE-LT (VLM $+$ TrEP) and GLANCE-LT (VLM $+$ PVIM).

\subsection{Cue-guided semantic stream}
\label{sec:semantic}
\noindent \textbf{Input.}
Eight of the 16 frames are sampled uniformly and tiled $4\times2$ into one image, the contact sheet, with the target in a red bbox in every frame. The box lets the model find the target without extra modules~\cite{10376723,10657559}, and the grid lets an image model read a short video~\cite{imagegrid}. The ego speed is given as one sentence, for example "Over the past 0.5 s the ego-vehicle decelerated from 23 km/h to 18 km/h.".
\begin{figure}[tb]
\centering\footnotesize
\fbox{\begin{minipage}{0.95\linewidth}
\textbf{System}. You are an autonomous vehicle safety system. Your task is
to predict whether a pedestrian will cross the road two to four seconds after the
last frame, based on visual evidence from a front-facing camera.
\end{minipage}}\\[3pt]
\fbox{\begin{minipage}{0.95\linewidth}
\textbf{Cue-guided prompt (\textbf{GLANCE-LT}).} [contact sheet] Analyse the pedestrian
marked with the red bounding box in the image sequence carefully. The question is
whether they cross two to four seconds after the last frame, not in the next
moment. \dots{} What shows instead is planning: they look at the road, wait for a
gap or a signal, and choose where to cross. Before answering, check each of the
following cues:\\[2pt]
\makebox[3.6em][l]{\textit{Person}}1.~\textsc{Head and gaze}: facing or looking at the road?\\
\makebox[3.6em][l]{}2.~\textsc{Waiting}: at the kerb, waiting for a gap or signal?\\
\makebox[3.6em][l]{}3.~\textsc{Movement}: towards the road, still, or along it?\\
\makebox[3.6em][l]{\textit{Scene}}4.~\textsc{Crossing place}: crosswalk, traffic light or intersection?\\
\makebox[3.6em][l]{}5.~\textsc{Traffic light}: red or green for the vehicle?\\
\makebox[3.6em][l]{\textit{Social}}6.~\textsc{Other people}: anyone else crossing or about to?\\
\makebox[3.6em][l]{\textit{Vehicle}}7.~\textsc{Ego-vehicle dynamics}: [speed sentence]\\[2pt]
Consider ALL of the above before deciding. \dots{} Answer with exactly one word
on the final line: yes or no.
\end{minipage}}
\caption{The cue-guided prompt, shortened (full text in the supplementary material). The no-cue prompt of the ablation keeps only the system text, the contact sheet and the question.}
\label{fig:prompt}
\end{figure}

\noindent \textbf{Guiding the model with the long-term cues.}
A naive prompt only asks whether the pedestrian will cross and does not tell the model what evidence matters at this range. Our prompt (Fig.~\ref{fig:prompt}) passes on the evidence of Section~\ref{sec:analysis}: it states that the question concerns two to four seconds ahead, explains that a crossing then shows in planning rather than in movement, and asks the model to check seven cues. The cues follow the groups of Section~\ref{sec:analysis}: the person (R1), the scene and the people around (R2), and the vehicle. The model then answers with a single word, yes or no. Each cue or group can be removed from the prompt without changing the rest, which gives the ablation of Section~\ref{sec:ablation_study}. With all cues removed, only the naive question remains: will the pedestrian cross two to four seconds after the last frame?

\noindent \textbf{Adaptation.}
Guidance tells the model what to look for, but it does not teach it the task. We therefore adapt the model with LoRA~\cite{hu2022lora}, which keeps the base weights frozen and trains small low-rank updates, on long-term windows ($\tau\in [60,120]$) with the same prompt and the correct answer as the target. Every result of the semantic stream, alone or fused, uses this adapted model.

\noindent \textbf{Score.}
Instead of parsing the generated word, we read the model's output distribution at the answer position, sum the probability of the token variants of yes and of no and normalise: $q_{\text{sem}} = P(\text{yes}) / (P(\text{yes}) + P(\text{no}))$.
This continuous score can be thresholded and fused like the geometric one.
\subsection{Implementation details}
\label{sec:impl}
The semantic stream is Qwen2.5-VL-32B-Instruct~\cite{bai2025qwen25vl} in bfloat16 on one NVIDIA H100 (93 GB). In the contact sheet each frame is resized to $336\times336$ px and the eight frames are tiled into one $1344\times672$ image. LoRA uses rank 8, $\alpha=16$ and dropout 0.1 on all linear layers. We train for one epoch with AdamW, a learning rate of $1\times10^{-4}$, weight decay 0.01, a cosine schedule with 5\% warm-up, gradient clipping at norm 1.0 and a class-balanced loss. The adapter is trained once with seed 0 and every ablation run uses the same configuration and seed. TrEP and PVIM use the authors' released code and hyper-parameters, trained on $\tau\in[60,120]$. TrEP is trained for three seeds and reported as the mean. The pose keypoints for
PVIM are those released with the PCPA benchmark~\cite{kotseruba2021pcpa}.

\begin{table*}[t]
\centering
\footnotesize
\renewcommand{\arraystretch}{0.92}
\setlength{\tabcolsep}{2.9pt}
\begin{tabular}{l ccc ccc ccc ccc ccc}
\toprule
\multicolumn{4}{l}{\textbf{Test set: PIE set03}} & \multicolumn{3}{c}{\textit{4 s}} & \multicolumn{3}{c}{\textit{3 s}} & \multicolumn{3}{c}{\textit{2 s}} & \multicolumn{3}{c}{\textit{1 s}} \\
\cmidrule(lr){5-7}\cmidrule(lr){8-10}\cmidrule(lr){11-13}\cmidrule(lr){14-16}
Method & Weights & Obs. & TTE range & AUC & F1 & Acc & AUC & F1 & Acc & AUC & F1 & Acc & AUC & F1 & Acc \\
\midrule
TrEP~\cite{zhang2023trep} (2023) & retrained & 16 frames & 1--2 s & 0.83 & 0.62 & 0.78 & 0.87 & 0.68 & 0.81 & 0.88 & 0.72 & 0.84 & 0.87 & 0.75 & \textbf{0.87} \\
VMI~\cite{sharma2023visualmotioninteraction} (2023) & retrained & 16 frames & 2.5/4 s & 0.70 & 0.50 & 0.62 & 0.69 & 0.50 & 0.59 & 0.72 & 0.53 & 0.66 & 0.67 & 0.49 & 0.57 \\
TAMformer~\cite{osman2023tamformer} (2023) & self-reported & 16--106 frames & 0.1--4 s & 0.77 & 0.65 & 0.78 & 0.81 & 0.70 & 0.81 & 0.84 & 0.76 & 0.87 & 0.88 & 0.80 & 0.88 \\
TAMformer~\cite{osman2023tamformer} (2023) & retrained & 16 frames & 0.1--4 s & 0.49 & 0.43 & 0.27 & 0.68 & 0.43 & 0.28 & 0.45 & 0.44 & 0.28 & 0.50 & 0.44 & 0.28 \\
TAMformer~\cite{osman2023tamformer} (2023) & second retrained & 16 frames & 0.1--4 s & 0.74 & 0.60 & 0.74 & 0.78 & 0.63 & 0.77 & 0.81 & 0.64 & 0.78 & 0.50 & 0.44 & 0.28 \\
CIPF~\cite{10208965} (2023) & self-reported & 15 frames & 1--4 s & 0.74 & 0.61 & 0.78 & 0.80 & 0.70 & 0.80 & 0.83 & 0.74 & 0.84 & 0.89 & 0.84 & 0.91 \\
PVIM~\cite{xu2024pvim} (2024) & retrained & 16 frames & 1--2 s & 0.82 & 0.64 & 0.77 & 0.84 & 0.66 & 0.78 & 0.88 & \textbf{0.74} & 0.84 & 0.89 & 0.73 & 0.84 \\
ViCross~\cite{tian2026vicross} (2027) & released & 16 frames & 1--2 s & 0.71 & 0.48 & 0.64 & 0.73 & 0.49 & 0.65 & 0.74 & 0.56 & 0.76 & 0.74 & 0.54 & 0.72 \\
ViCross~\cite{tian2026vicross} (2027) & retrained & 16 frames & 1--2 s & 0.58 & 0.32 & 0.67 & 0.58 & 0.31 & 0.67 & 0.61 & 0.40 & 0.71 & 0.62 & 0.39 & 0.69 \\
\hdashline\noalign{\vskip 2pt}
GLANCE-LT (VLM only) & ours & 16 frames & 2--4 s & 0.78 & 0.55 & 0.77 & 0.83 & 0.58 & 0.80 & 0.86 & 0.63 & 0.81 & 0.85 & 0.60 & 0.80 \\
GLANCE-LT (VLM $+$ PVIM) & ours & 16 frames & 2--4 s & 0.83 & 0.64 & 0.78 & 0.86 & 0.70 & 0.82 & \textbf{0.89} & \textbf{0.74} & \textbf{0.85} & \textbf{0.90} & 0.75 & 0.86 \\
GLANCE-LT (VLM $+$ TrEP) & ours & 16 frames & 2--4 s & \textbf{0.84} & \textbf{0.66} & \textbf{0.81} & \textbf{0.89} & \textbf{0.73} & \textbf{0.84} & \textbf{0.89} & \textbf{0.74} & \textbf{0.85} & 0.88 & \textbf{0.78} & \textbf{0.87} \\
\bottomrule
\end{tabular}
\caption{Results on PIE dataset~\cite{rasouli2019pie} set03, mean over 20 splits. Self-reported: numbers from the authors' paper, under their own protocol. Released: the authors' weights. Retrained: retrained on 2--4 s with the authors' code. Second retrained: TAMformer retrained in a second setting, in which each output head only learns from the windows at its own TTE. Bold: best under our protocol. The inputs of each method are in the supplementary material.}
\label{tab:pie}
\end{table*}

\begin{table*}[t]
\centering
\footnotesize
\renewcommand{\arraystretch}{0.92}
\setlength{\tabcolsep}{2.9pt}
\begin{tabular}{l ccc ccc ccc ccc ccc}
\toprule
\multicolumn{4}{l}{\textbf{Test set: JAAD\textsubscript{beh + VGGT}, trained on PIE}} & \multicolumn{3}{c}{\textit{4 s}} & \multicolumn{3}{c}{\textit{3 s}} & \multicolumn{3}{c}{\textit{2 s}} & \multicolumn{3}{c}{\textit{1 s}} \\
\cmidrule(lr){5-7}\cmidrule(lr){8-10}\cmidrule(lr){11-13}\cmidrule(lr){14-16}
Method & Weights & Obs. & TTE range & AUC & F1 & Acc & AUC & F1 & Acc & AUC & F1 & Acc & AUC & F1 & Acc \\
\midrule
TrEP~\cite{zhang2023trep} (2023) & retrained & 16 frames & 1--2 s & 0.47 & 0.46 & 0.36 & 0.44 & \textbf{0.55} & 0.39 & 0.58 & 0.61 & 0.49 & 0.61 & 0.63 & 0.51 \\
VMI~\cite{sharma2023visualmotioninteraction} (2023) & retrained & 16 frames & 2.5/4 s & 0.56 & 0.44 & 0.43 & 0.49 & 0.51 & 0.41 & 0.47 & 0.56 & 0.42 & 0.58 & 0.62 & 0.48 \\
TAMformer~\cite{osman2023tamformer} (2023) & retrained & 16 frames & 0.1--4 s & 0.27 & 0.50 & 0.33 & 0.46 & 0.57 & 0.40 & 0.58 & \textbf{0.62} & 0.44 & 0.50 & 0.67 & 0.50 \\
TAMformer~\cite{osman2023tamformer} (2023) & second retrained & 16 frames & 0.1--4 s & 0.38 & 0.44 & 0.34 & 0.48 & \textbf{0.55} & 0.39 & 0.47 & \textbf{0.62} & 0.45 & 0.50 & 0.67 & 0.50 \\
PVIM~\cite{xu2024pvim} (2024) & retrained & 16 frames & 1--2 s & 0.55 & 0.44 & 0.44 & 0.48 & \textbf{0.55} & 0.41 & 0.59 & 0.61 & 0.51 & 0.58 & 0.66 & 0.55 \\
ViCross~\cite{tian2026vicross} (2027) & released & 16 frames & 1--2 s & 0.48 & \textbf{0.47} & 0.38 & 0.35 & 0.52 & 0.37 & 0.38 & \textbf{0.62} & 0.45 & 0.44 & 0.65 & 0.51 \\
ViCross~\cite{tian2026vicross} (2027) & retrained & 16 frames & 1--2 s & 0.48 & 0.26 & \textbf{0.70} & 0.42 & 0.09 & 0.59 & 0.36 & 0.09 & 0.50 & 0.42 & 0.06 & 0.46 \\
\hdashline\noalign{\vskip 2pt}
GLANCE-LT (VLM only) & ours & 16 frames & 2--4 s & \textbf{0.63} & 0.44 & 0.65 & \textbf{0.57} & 0.45 & \textbf{0.64} & \textbf{0.66} & 0.59 & \textbf{0.70} & 0.60 & 0.53 & 0.60 \\
GLANCE-LT (VLM $+$ PVIM) & ours & 16 frames & 2--4 s & 0.60 & 0.40 & 0.53 & 0.56 & 0.47 & 0.47 & \textbf{0.66} & 0.58 & 0.59 & 0.61 & \textbf{0.68} & 0.57 \\
GLANCE-LT (VLM $+$ TrEP) & ours & 16 frames & 2--4 s & 0.56 & \textbf{0.47} & 0.57 & \textbf{0.57} & 0.47 & 0.52 & \textbf{0.66} & 0.57 & 0.58 & \textbf{0.64} & 0.67 & \textbf{0.61} \\
\bottomrule
\end{tabular}
\caption{Generalisation test on JAAD\textsubscript{beh}~\cite{kotseruba2016jaad}, with every model trained on PIE only. JAAD~\cite{kotseruba2016jaad} has no ego speed, so it is estimated from the camera with VGGT~\cite{wang2025vggt}. Mean over 20 splits. Weights as in Table~\ref{tab:pie}. Bold: best per column.}
\label{tab:jaad}
\end{table*}

\section{Experiments}
\label{sec:setup}
\subsection{Experimental setup}
\noindent \textbf{Datasets.}
\label{sec:datasets}
We evaluate all methods on PIE~\cite{rasouli2019pie}, our main dataset and on JAAD~\cite{kotseruba2016jaad} as a robustness check (Section~\ref{sec:related_datasets}). On JAAD we use the pedestrians with behaviour labels ($JAAD\textsubscript{beh}$) and the test videos of the default split of Kotseruba~\etal~\cite{kotseruba2016jaad}. 

\noindent \textbf{Protocol and metrics.}
\label{sec:protocol}
Every model receives the same 16-frame window (Section~\ref{sec:task}) and is tested at a TTE of 1, 2, 3, and 4 s. Every trained model, ours and the baselines, is trained on PIE set01, set02, and set04 with a TTE of 2 to 4 s. Set05 and set06 are used for validation and set03 for testing. JAAD is only used for testing with the ego speed estimated from the camera with VGGT~\cite{wang2025vggt}. Instead of a fixed threshold of 0.5, the threshold $\theta$ of every method and for a fusion also the weight $\beta$ is tuned for the best F1 on one stratified half of the test set and the method is scored on the other half. We repeat this for 20 splits and report the mean AUC, F1 and accuracy. Details are in the supplementary material.

\noindent \textbf{Baselines.}
\label{sec:baselines}
We compare against five published methods whose code is public and works with minimal adjustments (Section~\ref{sec:related_pcp}): TrEP~\cite{zhang2023trep}, VMI~\cite{sharma2023visualmotioninteraction}, TAMformer~\cite{osman2023tamformer}, PVIM~\cite{xu2024pvim}, and ViCross~\cite{tian2026vicross}. Table~\ref{tab:pie} also lists the self-reported results of TAMformer and CIPF~\cite{10208965}, which both report up to 4 s on PIE, and ViCross with its released weights.
 PVIM~\cite{xu2024pvim} and TrEP~\cite{zhang2023trep} are also our geometric streams. TAMformer is the only one designed for a TTE up to 4 s, and ViCross~\cite{tian2026vicross} is the closest to our approach, since it fine-tunes a VLM (Qwen2-VL-7B~\cite{yang2024qwen2technicalreport}).

Each baseline is retrained with the authors' code and hyper-parameters on our training windows, with three changes to keep the comparison fair. TAMformer predicts from the whole track, so we restrict its attention mask to the last 16 frames. TAMformer is very sensitive to how it is trained, so we also report it in a second setting, in which each output head only learns from the windows at its own TTE (second retrained in Tables~\ref{tab:pie} and~\ref{tab:jaad}). It improves clearly in this setting, but on PIE it stays below GLANCE-LT (VLM $+$ TrEP) at every TTE from 2 to 4 s. VMI's public code uses a random split, a single TTE of 2 s and the training set for validation, which we replace with our splits, windows and validation set. ViCross answers in text, so we take the crossing probability of its classification head as its score, which lets its threshold be tuned like ours.

\subsection{Comparison on PIE datasets}

\label{sec:results_pie}

Table~\ref{tab:pie} compares all methods on PIE~\cite{rasouli2019pie} set03. Three results stand out. First, GLANCE-LT (VLM $+$ TrEP) achieves the best overall performance in the long term. It gives the best AUC, F1 and accuracy at 3 and 4 s. At 4 s it reaches an F1 of 0.66 against 0.64 for PVIM~\cite{xu2024pvim}, the best baseline. At 3 s it is 0.73 against 0.68 for TrEP~\cite{zhang2023trep}, the best baseline. At 2 s it ties with PVIM~\cite{xu2024pvim} on F1 (0.74) and is better on AUC and accuracy. At 1 s, outside its training range, it still gives the best F1 under our protocol (0.78 against 0.75 for TrEP~\cite{zhang2023trep}).

Second, using a VLM alone we already outperform the existing VLM only methods (ViCross~\cite{tian2026vicross}). We compare with two versions of ViCross: the released weights, which are optimised for short-term prediction, and a retrained version for the long term. Our GLANCE-LT (VLM only) outperforms both versions at every TTE. At 3 s, for example, GLANCE-LT (VLM only) reaches an AUC of 0.83, which is 0.10 higher than the released ViCross (0.73) and 0.25 higher than the retrained one (0.58). 

Third, the geometric models TrEP~\cite{zhang2023trep} and PVIM~\cite{xu2024pvim} alone outperform GLANCE-LT (VLM only) in AUC and F1 on PIE. On JAAD, however, their AUC falls to around chance at 3 and 4 s (Section~\ref{sec:results_jaad}), which indicates that they may overfit PIE and lack generalisation.

\begin{figure*}[t]
\centering
\includegraphics[width=\textwidth]{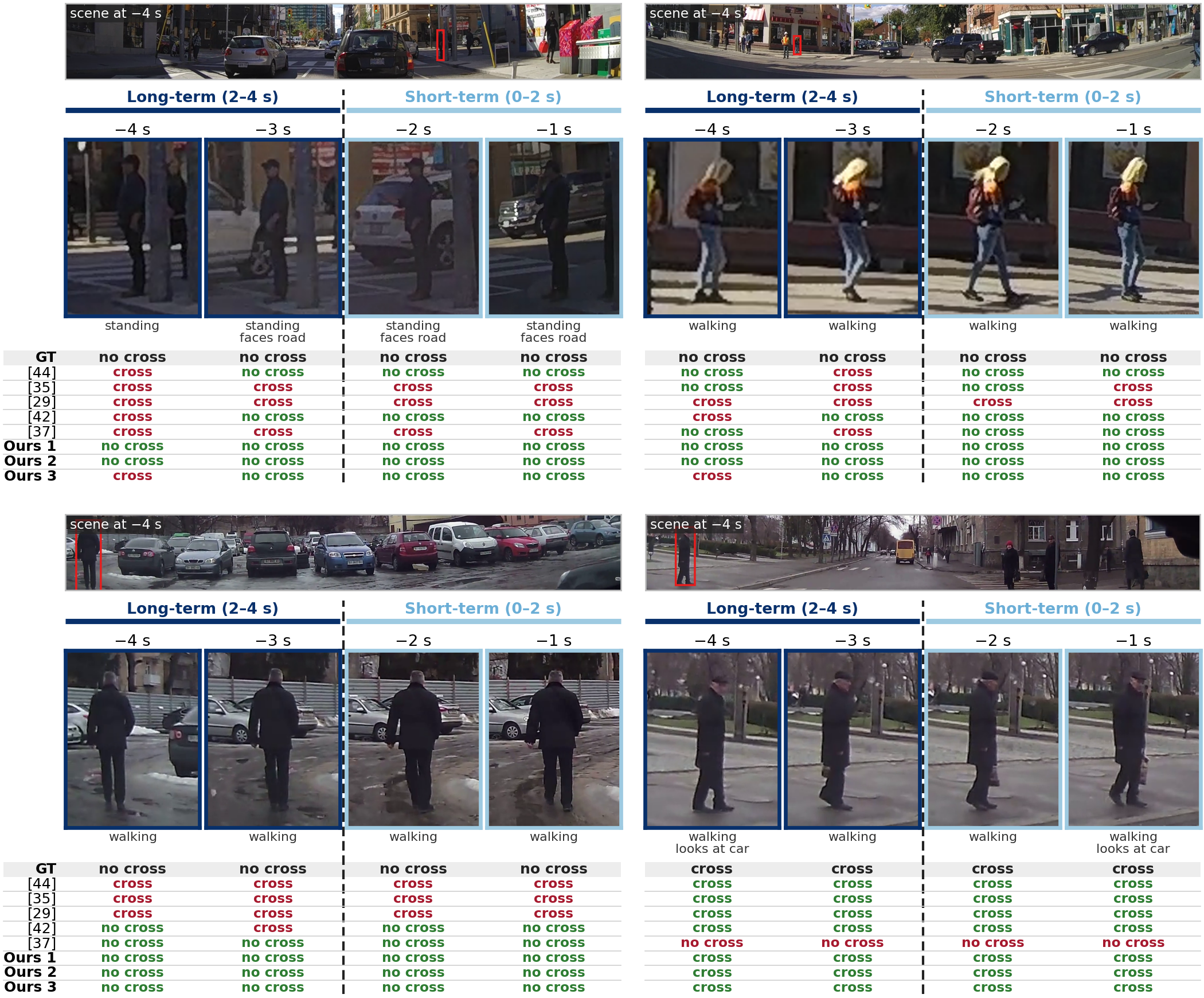}
\caption{Long-term ($>$2 s) vs short-term ($\leq$2 s) pedestrian crossing prediction on PIE~\cite{rasouli2019pie} (top) and JAAD~\cite{kotseruba2016jaad} (bottom, without training on JAAD). Top: two non-crossers. Bottom left: a non-crosser. Bottom right: a crosser. Our methods model human behaviours related to long-term planning and make the correct prediction at almost every moment, while every baseline is wrong at least once on both PIE pedestrians. Compared methods: TrEP (2023)~\cite{zhang2023trep}, VMI (2023)~\cite{sharma2023visualmotioninteraction}, TAMformer (2023)~\cite{osman2023tamformer}, PVIM (2024)~\cite{xu2024pvim} and ViCross~\cite{tian2026vicross}. Ours 1: VLM only, Ours 2: VLM + TrEP, Ours 3: VLM + PVIM.}

\label{fig:visual}
\end{figure*}
\subsection{Comparison on JAAD datasets}
\label{sec:results_jaad}
Table~\ref{tab:jaad} shows the transfer from PIE~\cite{rasouli2019pie} to JAAD~\cite{kotseruba2016jaad}, to test whether the methods generalise. Every model is trained on PIE only, and only the threshold and $\beta$ are set on JAAD, on one half of each split. JAAD has more crossers than PIE, 34\% of the windows at 4 s and 50\% at 1 s against 27\% and 28\% on PIE.

The geometric models do not transfer in the long-term range. At 3 and 4 s, the AUC of TrEP~\cite{zhang2023trep}, PVIM~\cite{xu2024pvim}, and VMI~\cite{sharma2023visualmotioninteraction} stays between 0.44 and 0.56, around chance, TAMformer~\cite{osman2023tamformer} is at or below chance, and ViCross~\cite{tian2026vicross} with its released weights reaches 0.35 and 0.48. Retrained on two to four seconds, ViCross~\cite{tian2026vicross} falls below chance at every TTE (0.36 to 0.48) and predicts a crossing for almost no pedestrians, so its accuracy of 0.70 at 4 s, comes from the non-crossers, who make up 66\% of the windows.
Closer to the event TrEP and PVIM recover, to an AUC of 0.58 and 0.59 at 2 s. GLANCE-LT (VLM only) stays between 0.57 and 0.66 at every TTE. This is consistent with Section~\ref{sec:analysis}, where moving towards the road only separates crossers from non-crossers in the last two seconds. The estimated speed itself is unlikely to be the cause, since TrEP scores the same with the VGGT and the OBD speed on PIE (supplementary material).

\subsection{Visual results}
Fig.~\ref{fig:visual} shows two non-crossers from PIE~\cite{rasouli2019pie} and two pedestrians from JAAD~\cite{kotseruba2016jaad}, with the ground truth and whether each method is right at each TTE. The first PIE non-crosser waits at a crosswalk and turns the head to the road. Every baseline predicts a crossing at $-4$ s, while GLANCE-LT (VLM only) and GLANCE-LT (VLM $+$ TrEP) are right throughout. The second walks along the pavement, where again every baseline is wrong at least once and these two models are right. On JAAD, without training on it, TrEP~\cite{zhang2023trep}, VMI~\cite{sharma2023visualmotioninteraction}, and TAMformer~\cite{osman2023tamformer} predict a crossing at every moment for the non-crosser who walks through a car park, while all our models are right throughout. Every method except ViCross~\cite{tian2026vicross} predicts the crosser who walks to the crosswalk and looks at the car.

\subsection{Ablation study}
\label{sec:ablation_study}
\noindent \textbf{How useful are the proposed cues?}
We examine the usage of the proposed seven cues (Fig.~\ref{fig:prompt}),
 by leaving out one of the seven cues at a time and by adding one cue at a time to a prompt without cues. Every run retrains the LoRA~\cite{hu2022lora} adapter with the configuration and seed of the final model and is fused with TrEP~\cite{zhang2023trep} and PVIM~\cite{xu2024pvim} under the protocol of Table~\ref{tab:pie}.
\\\\
Every cue contributes on its own (Table~\ref{tab:ablation_abs}). Without cues GLANCE-LT (VLM only) scores an AUC of 0.60 and an F1 of 0.31 at 4 s. A single cue raises both at every TTE, except crossing place, which lowers AUC at 4 s to 0.59. Ego-vehicle speed raises F1 at 4 s the most, to 0.56, followed by traffic light (0.53) and head and gaze (0.50).

Leaving out one cue costs less, because the cues overlap: waiting and movement both ask whether the pedestrian stands still, and a visible traffic light usually means a crossing place. Only ego-vehicle speed is hard to replace: without it, F1 at 4 s falls from 0.55 to 0.38 and AUC from 0.78 to 0.70. Without movement, AUC falls at 2 and 1 s (0.86 to 0.84 and 0.85 to 0.82), matching Section~\ref{sec:analysis}, and without other people AUC is lower at every TTE. After fusion the cues still matter: without any cue, GLANCE-LT (VLM $+$ TrEP) reaches an F1 of 0.62 at 4 s, the same as TrEP alone, against 0.66 with all cues (supplementary material).

\noindent \textbf{How useful is the geometric stream?} In Tables~\ref{tab:pie} and \ref{tab:jaad}, we  test two implementations of the geometric stream, TrEP~\cite{zhang2023trep} and PVIM~\cite{xu2024pvim}. On their own they perform similarly on PIE~\cite{rasouli2019pie}, with an F1 of 0.62 to 0.75 for TrEP and 0.64 to 0.74 for PVIM. Nevertheless, adding the semantic stream never lowers the F1 of either, and it raises it most at 3 s. The streams of GLANCE-LT are therefore complementary. At 4 s TrEP alone reaches an F1 of 0.62 and GLANCE-LT (VLM only) 0.55, and together 0.66, and at 3 s 0.68 and 0.58 become 0.73. With PVIM the gain is largest at 3 s, where F1 rises from 0.66 for PVIM alone to 0.70. The semantic stream therefore adds evidence that the geometric models do not have and on PIE~\cite{rasouli2019pie} the best results come from both streams together.

\noindent \textbf{Choice of VLM.}
We also replace Qwen2.5-VL-32B~\cite{bai2025qwen25vl} with Qwen3-VL-8B~\cite{bai2025qwen3vltechnicalreport}, a VLM four times smaller, with the same LoRA configuration, prompt and contact sheets. GLANCE-LT (VLM $+$ TrEP) then stays within 0.02 F1 of the 32B model at 2 to 4 s (supplementary material).

\begin{table}[t]
\centering
\resizebox{\columnwidth}{!}{%
\begin{tabular}{@{}lcccccccc@{}}
\toprule
 & \multicolumn{2}{c}{4\,s} & \multicolumn{2}{c}{3\,s} & \multicolumn{2}{c}{2\,s} & \multicolumn{2}{c}{1\,s} \\
\cmidrule(lr){2-3}\cmidrule(lr){4-5}\cmidrule(lr){6-7}\cmidrule(lr){8-9}
Prompt & F1 & AUC & F1 & AUC & F1 & AUC & F1 & AUC \\
\midrule

\multicolumn{9}{@{}l}{\textit{Leave one out}} \\
All cues & 0.55 & 0.78 & 0.58 & 0.83 & 0.63 & 0.86 & 0.60 & 0.85 \\
Without head and gaze & 0.55 & 0.79 & 0.61 & 0.85 & 0.61 & 0.86 & 0.65 & 0.84 \\
Without waiting & 0.53 & 0.78 & 0.59 & 0.84 & 0.62 & 0.85 & 0.60 & 0.82 \\
Without movement & 0.52 & 0.80 & 0.57 & 0.83 & 0.59 & 0.84 & 0.58 & 0.82 \\
Without crossing place & 0.55 & 0.76 & 0.58 & 0.81 & 0.67 & 0.85 & 0.69 & 0.85 \\
Without traffic light & 0.55 & 0.76 & 0.59 & 0.82 & 0.63 & 0.85 & 0.61 & 0.84 \\
Without other people & 0.53 & 0.76 & 0.59 & 0.81 & 0.60 & 0.83 & 0.56 & 0.82 \\
Without ego-vehicle speed & 0.38 & 0.70 & 0.44 & 0.73 & 0.47 & 0.75 & 0.44 & 0.74 \\
\midrule
\multicolumn{9}{@{}l}{\textit{One cue only}} \\
No cues & 0.31 & 0.60 & 0.37 & 0.59 & 0.44 & 0.64 & 0.41 & 0.65 \\
Only head and gaze & 0.50 & 0.72 & 0.55 & 0.75 & 0.58 & 0.78 & 0.56 & 0.73 \\
Only waiting & 0.49 & 0.72 & 0.52 & 0.75 & 0.57 & 0.78 & 0.51 & 0.71 \\
Only movement & 0.47 & 0.75 & 0.53 & 0.77 & 0.50 & 0.80 & 0.51 & 0.72 \\
Only crossing place & 0.34 & 0.59 & 0.48 & 0.65 & 0.49 & 0.71 & 0.46 & 0.70 \\
Only traffic light & 0.53 & 0.77 & 0.56 & 0.77 & 0.59 & 0.80 & 0.53 & 0.73 \\
Only other people & 0.43 & 0.71 & 0.57 & 0.74 & 0.56 & 0.78 & 0.52 & 0.74 \\
Only ego-vehicle speed & 0.56 & 0.75 & 0.62 & 0.81 & 0.70 & 0.87 & 0.66 & 0.84 \\

\bottomrule
\end{tabular}}
\caption{Cue ablation of GLANCE-LT (VLM only) on PIE set03.}
\label{tab:ablation_abs}
\end{table}

\subsection{Limitations}
\label{sec:limitations}
Two restrictions limited how far we could test long-term prediction: GPU memory and dataset size.

To fit the 32B model on one H100, each frame is resized to $336\times336$ px, so a distant pedestrian covers only a few pixels (R3), and the adapter is trained for one epoch with one seed. Within these limits the semantic stream alone (GLANCE-LT (VLM only)) scores below TrEP~\cite{zhang2023trep} and PVIM~\cite{xu2024pvim} alone on PIE~\cite{rasouli2019pie} (Table~\ref{tab:pie}). The geometric models already use all the input that PIE~\cite{rasouli2019pie} annotates, while the semantic stream can still gain from more data, more training runs and more resolution, such as a crop of the pedestrian next to the full frame.
The datasets also limit further study. The tracks of PIE and JAAD are short: the PIE test windows fall from 694 at 1 s to 508 at 4 s, JAAD has only 59 at 4 s, and PIE was recorded in one city. A study beyond 4 s is therefore not possible on these datasets, and new datasets with long tracks from several cities are needed.
\\
\section{Conclusion and future work}
\label{sec:conclusion}
In this work, we investigated long-term pedestrian crossing prediction and demonstrated the importance of reasoning about human planning beyond observable motion. We proposed GLANCE-LT, a planning-aware framework that combines cue-guided vision-language modeling with geometric reasoning to predict crossing intentions two to four seconds in advance. Extensive experiments on PIE demonstrate state-of-the-art performance, while cross-dataset evaluation on JAAD highlights the generalization capability of the semantic stream. Ablation studies further confirm the effectiveness of planning-aware cues and the complementarity of semantic and geometric information. Our findings establish human planning-aware reasoning as a promising direction for long-term pedestrian crossing prediction. For future work, we aim for long-term prediction beyond 4 s and new datasets.

\clearpage
\section*{LLM Usage Statement}
The authors used Claude (Anthropic), specifically Claude Fable 5.2 and Claude Opus 5.5, as an assistant in two ways. The first is to organize experiments and write Slurm jobs, where it helped with the runs and helped write and debug code. For the paper, it helped to polish English prose and present the figures. The authors defined the research direction and designed the experiments and take responsibility for the method, citations, results, and the final manuscript.

{
    \small
    \bibliographystyle{ieeenat_fullname}
    \bibliography{main}
}

\end{document}

%% file: preamble.tex
\usepackage{xcolor}
\usepackage{tikz}
\usepackage{enumitem}
\definecolor{barblue}{HTML}{2A78D6}
\usepackage{float}
\usepackage{caption}
\usepackage{capt-of}
\usepackage{arydshln}
\usepackage{pgfplots}
\pgfplotsset{compat=1.18}

